\documentclass[10pt]{article}
\usepackage{graphics,epsfig}
\usepackage{harvard}
\usepackage{amsmath,amssymb,amsthm}
\usepackage{graphicx}
\usepackage[tocindex]{apacite}
\usepackage[font=small,labelfont=bf]{caption}
\usepackage{url}

\newcommand{\bibnodot}[1]{}

\begin{document}
\noindent
{\bf Quantum and Concept Combination, Entangled Measurements and Prototype Theory}

\bigskip
\noindent
Diederik Aerts

\noindent
Center Leo Apostel for Interdisciplinary Studies

\noindent
Brussels Free University, Brussels, Belgium

\noindent
E-Mail: \textsf{diraerts@vub.ac.be}

\bigskip
\noindent
\small
{\bf Abstract:} We analyze the meaning of the violation of the marginal probability law for situations of correlation measurements where entanglement is identified. We show that for quantum theory applied to the cognitive realm such a violation does not lead to the type of problems commonly believed to occur in situations of quantum theory applied to the physical realm. We briefly situate our quantum approach for modeling concepts and their combinations with respect to the notions of `extension' and `intension' in theories of meaning, and in existing concept theories.
\normalsize

\bigskip
\noindent
This article is a reply to the publications Dzhafarov \& Kujala (2013) and Hampton (2013), which are commentaries on Aerts, Gabora \& Sozzo (2013), and contains at the same time specifications and elaborations of many aspects touched upon in Aerts, Gabora \& Sozzo (2013) and earlier publications (Aerts, 2009; Aerts \& Gabora, 2005; Aerts \& Sozzo, 2011) on the same topic.

Dzhafarov \& Kujala (2013)'s commentary focuses on the section about `entanglement' in Aerts, Gabora \& Sozzo (2013), where we present an example of two concepts and their combination, and prove that well chosen coincidence experiments by human subjects on this combination of concepts produce data that violate the Clauser, Horne Shimony and Holt (CHSH) inequality, which is a version of Bell's inequality. We presented this example originally in Aerts \& Sozzo (2011), where it is worked out in more detail, and more recently elaborated a complete quantum complex Hilbert space representation (Aerts \& Sozzo, 2013a). This more recent material was not known to Dzhafarov \& Kujala (2013) when they wrote their commentary, otherwise they would have seen that their critique on our example is substantially accounted for by the identification of, next to entangled states, the presence of entangled measurements representing the coincidence experiments. This result, proving the need for `entangled measurements' to represent the experimental data, is subtle and not straightforward -- we have meanwhile scrutinized extensively the literature in quantum foundations physics, and found that the notion of `entangled measurement' has not been studied at all. As we will show in the following, although we do not agree with Dzhafarov \& Kujala (2013)Õs critique, we find the material in their article and their associated work very valuable and interesting, because it touches upon these subtle matters directly, and also gives us the opportunity to respond to it and clarify issues in our own work. Let us first formulate the necessary notions.

To formulate the CHSH inequality, one considers experiments $e_{A}$ and $e_{A'}$ on system $S_{A}$ and experiments $e_{B}$ and $e_{B'}$ on system $S_{B}$, each with two possible outcomes, `1' and `2', and coincidence experiments $e_{AB}$, $e_{AB'}$, $e_{A'B}$ and $e_{A'B'}$ on system $S_{AB}$, joint system of $S_{A}$ and $S_{B}$, each with four possible outcomes being the combinations of outcomes `1' and `2', hence `11', `12', `21' and `22'. We denote the probabilities for the respective outcomes, for experiment $e_{A}$, $\{p(A_1), p(A_2)\}$, for experiment $e_{A'}$, $\{p(A'_1), p(A'_2)\}$, for experiment $e_{B}$, $\{p(B_1), p(B_2)\}$, for experiment $e_{B'}$, $\{p(B'_1), p(B'_2)\}$, for experiment $e_{AB}$, $\{p(A_1B_1), p(A_1B_2), p(A_2B_1), p(A_2B_2)\}$, for experiment $e_{AB'}$, $\{p(A_1B'_1), p(A_1B'_2), p(A_2B'_1), p(A_2B'_2)\}$, for experiment $e_{A'B}$, $\{p(A'_1B_1), p(A'_1B_2), p(A'_2B_1), p(A'_2B_2)\}$, and for experiment $e_{A'B'}$, $\{p(A'_1B'_1), p(A'_1B'_2), p(A'_2B'_1),$ $p(A'_2B'_2)\}$. Hence, for example, $p(A'_1B_2)$ is the probability that the coincidence experiment $e_{A'B}$ gives outcome `12'. The expectations values are defined as usual, for example $E(A)=p(A_1)-p(A_2)$, and $E(A,B)=p(A_1,B_1)-p(A_1,B_2)-p(A_2,B_1)+p(A_2,B_2)$. The CHSH inequality is the following $-2 \le E(A',B')+E(A',B)+E(A,B')-E(A,B)\le 2$.

Dzhafarov \& Kujala (2013)Õs critique of our entanglement example is that `we only looked at the violation of the CHSH inequality, and not paid attention to another requirement -- which they call `marginal selectivity', i.e. equation (4) in Dzhafarov \& Kujala (2013), and which we, in Aerts \& Sozzo (2013a), have indicated with `the marginal distribution law'. Let us explain what this `marginal distribution law' is, because it is a core element for our further analysis.

In the experimental setting corresponding to CHSH as described above one can focus specifically on the probabilities obtained by considering only one of the subsystems, while neglecting what happens on the other subsystem. For example, we can consider the situation to find outcome `1' related to $A$ for subsystem $S_{A}$, in the three following cases, (i) when this outcome is obtained by executing $e_{A}$, the corresponding probability is $p(A_1)$, (ii) when this outcome is obtained by executing $e_{AB}$, the corresponding probability is $p(A_1B_1)+p(A_1B_2)$ (iii) when this outcome is obtained by executing $e_{AB'}$, the corresponding probability is $p(A_1B'_1)+p(A_1B'_2)$. Let us consider now the experimental data for our concrete experiment on {\it Animal}, {\it Acts} and their combination {\it The Animal Acts} (Aerts \& Sozzo, 2011, 2013a). For $e_{A}$ subjects were asked to choose between {\it Horse} (outcome `1'), and {\it Bear} (outcome `2'), for {\it Animal}. We found $p(A_1)=0.531$. For $e_{AB}$ subjects were asked to choose between {\it The Horse Growls} (outcome `11'), {\it The Horse Whinnies} (outcome `12'), {\it The Bear Growls} (outcome `21'), or {\it The Bear Whinnies} (outcome `22'), for {\it The Animal Acts}. We found $p(A_1,B_1)+p(A_1,B_2)=0.049+0.630=0.679$. For $e_{AB'}$ subjects were asked to choose between {\it The Horse Snorts} (outcome `11'), {\it The Horse Meows} (outcome `12'), {\it The Bear Snorts} (outcome `21') or {\it The Bear Meows} (outcome `22'), for {\it The Animal Acts}. We found $p(A_1,B'_1)+p(A_1,B'_2)=0.593+0.025=0.618$. These three probabilities, 0.531, 0.679 and 0.618 are different. Subjects choose differently for {\it Horse} or {\it Bear} when no context of an animal sound is present (then {\it Horse} gets probability 0.531), or when the context $\{\it Growls, Whinnies\}$ as animal sound is present (then {\it Horse} gets probability 0.679), or when in the context $\{\it Snorts, Meows\}$ as animal sound is present (then {\it Horse} gets probability 0.618).

Satisfying the marginal distribution law -- or marginal selectivity (Dzhafarov \& Kujala, 2013) -- means that the last two of these probabilities are equal (in principle we can consider `no context' to be also a context, and hence the marginal distribution law would then imply all three probabilities to be equal). We now have all elements needed to consider in detail the critique of Dzhafarov \& Kujala (2013), and formulate our response together with a clear description of what we believe the situation to be. Along the view of Dzhafarov \& Kujala (2013), (i) the marginal probability law to be satisfied, and (ii) Bell's inequalities to be violated, are both considered necessary for one to be able to speak of `the identification of quantum entanglement'. Since in our example, the marginal probability law is violated, they do not consider it an example entailing genuine quantum entanglement. We do not agree with this view for several reasons, which will be listed and supported in the following.

1) The main reason to see a profound problem in the violation of the marginal distribution law in quantum theory applied to the physical realm, is because there is a specific problematic situation concerning the `collapse of the wave function' in quantum theory applied to the physical realm. Is this collapse instantaneous? Is there a physical process corresponding to this collapse? These are two questions that have no agreed upon answer in quantum theory applied to the physical realm. In quantum theory applied to cognition, hence in quantum cognition, the collapse does represent a `real change of state', and `is not instantaneous', which means that the two problematic questions are answered. For example, for the considered experiments, the collapse represents the choice made by the subjects in the experiment. This means that for quantum cognition `marginal selectivity' should not be treated as a property `at all means to be avoided to be violated', or as a property `equally important to consider as the violation of Bell's inequality when it comes to identifying entanglement'. As we have proven, `marginal selectivity' is equivalent to `the measurements being product measurements', hence `marginal selectivity' means that the situation contains less entanglement, and a greater degree of symmetry, such that all the entanglement can be concentrated in the state (Aerts \& Sozzo, 2013a).

2) Independently of 1), one can believe that `the situation with only entanglement in the state, and no entanglement in the measurements contains the genuine weirdness of quantum entanglement', namely `correlations without manifest influence between measurements'. We know that this is a general view in the community of quantum foundation physicists, and we believe it to be wrong for several reasons. We will make our argument by looking carefully at some examples of correlated systems. It is some time ago that we studied these matters, and these results are not well known in todayÕs community of quantum foundations physicists. Indeed, at the time they were obtained, not much interest existed in situations different from the micro-world entailing quantum structure, since most physicists were convinced that it was not possible for such situations to exist. Now, with quantum cognition flourishing, it is interesting to take a fresh look at them, since they shed light on the very question we are analyzing here. 

We proposed an example of a macroscopic non local system consisting of `vessels of water connected by a tube', and `violating Bell's inequality' (Aerts, 1982, 1983), and we will now describe this example and show that (i) it violates the marginal distribution law, (ii) produces correlations that take place at space-like separated events, (iii) produces no signaling that violates relativity theory, and (iv) sheds light on the question why the marginal distribution law violation might (or might not) be a problem in quantum theory applied to the physical realm. In Figure 1 we have represented system $S_{AB}$ consisting of two vessels $V_A$ and $V_B$ connected by a tube $T$ containing a total of 20 liters of transparent water. On the left side, the experiment $e_{A}$ consists in pouring out the water through a siphon $S_{A}$ into a reference vessel $R_{A}$, where the volume of collected water is measured. Outcome `1' occurs when the volume is more than 10 liters, and outcome `2' occurs when the volume is less than 10 liters.
\begin{figure}[h]
\centerline {\includegraphics[width=10cm]{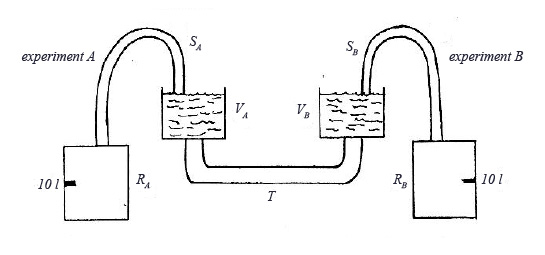}}
\caption{The system $S_{AB}$ consists of two vessels $V_A$ and $V_B$ connected by a tube $T$ containing a total of 20 liters of transparent water. Coincidence experiment $e_{AB}$ consist in siphons $S_A$ and $S_B$ pouring out water from vessels $V_A$ and $V_B$ and collecting the water in reference vessels $R_A$ and $R_B$, where the volume of collected water is measured.}
\end{figure}
Experiment $e_{A'}$ consists of taking a little spoon of water and verifying its transparency, outcome `1' occurs when transparent, and outcome `2', when not transparent. Analogous experiments are defined at the right vessels, hence $e_{B}$ and $e_{B'}$. Let us remark that the setup is an example of what in physics is called a system of `communicating vessels', and the way in which such system functions plays a role in our experiment. More specifically, the siphon experiments $e_{A}$ and $e_{B}$ will take out 20 liters of water, respectively to the left and to the right, only water in the tube remaining. Coincidence experiments $e_{AB}$, $e_{AB'}$, $e_{A'B}$ and $e_{A'B'}$ are defined as the joint experiments. For example, coincidence experiment $e_{AB}$ consist in siphons $S_A$ and $S_B$ pouring out water from vessels $V_A$ and $V_B$ and collecting the water in reference vessels $R_A$ and $R_B$, where the volume of collected water is measured. We only have to point out two obvious aspects of the experiment to derive the probabilities, and calculate the CHSH inequality. First, for coincidence experiment $e_{AB}$, again due to the setup being a system of communicating vessels, 20 liters of water are poured out in total, only water remaining in the tube. This means that if more than 10 liters are collected left, then fewer than 10 liters are collected right, and vice versa. Hence we have a perfect anti-correlation, i.e. $p(A_1,B_1)=p(A_2,B_2)=0$, $p(A_1,B_2)=p(A_2,B_1)={1 \over 2}$, and $E(A,B)=-1$. If a little spoonful of water is taken at one side, again due to the principles of communicating vessels, all of the remaining water will be poured out by the siphon at the other side. This means that always more than 10 liters will be collected at the other side, and hence for coincidence experiments $e_{AB'}$, $e_{A'B}$, since the water is transparent, we will always obtain the outcome `11'. Also for $e_{A'B'}$ we obtain always outcome `11'. Hence we have $p(A_1,B'_1)=p(A'_1,B_1)=p(A'_1,B'_1)=1$ and $p(A_1,B'_2)=p(A_2,B'_1)=p(A_2,B'_2)=p(A'_1,B_2)=p(A'_2,B_1)=p(A'_2,B_2)=p(A'_1,B'_2)=p(A'_2,B'_1)=p(A'_2,B'_2)=0$. Hence $E(A,B')=E(A',B)=E(A',B')=1$. This means that the CHSH expression becomes $E(A',B')+E(A,B')+E(A',B)-E(A,B)=4$, which gives rise to a maximal violation of the inequality. Let us verify the marginal distribution. We have $p(A_1,B_1)+p(A_1,B_2)={1 \over 2}$, and $p(A_1,B'_1)+p(A_1,B'_2)=1$, hence the marginal distribution law is violated too. In a forthcoming article we will give a full complex Hilbert space quantum description for this vessels of water system, and prove that entangled measurements are involved (Aerts \& Sozzo 2013b).

3) Let us analyze the example with the aim of understanding better what is involved in the violation of the marginal distribution law, and why quantum foundation physicists should (or should not) worry about this. First we remark the following. If we define the correlation events as the events when `the water stops flowing', then the correlation events of $e_{AB}$ are `space-like separated events', because indeed, due to the functioning of the setup as communicating vessels and the symmetry involved, the water stops flowing at both sides at the same time, in the reference frame of the apparatuses. Hence, if one event at one side influenced the other event at the other side (directly) through some mechanism, then this would involve `an influence faster than light'. Of course, the correlations are produced from the entangled state of the vessels -- entangled due to the connecting tube -- and then by means of the measurements pulling at this state on both sides to make it collapse. In other words, since there is no (simple) influence from the one side to the other, it makes no sense to speak of a velocity of such a (simple) influence. What we learn from this is that, `even if detection events (for example for quantum correlation of entangled photons) are space-like separated', this should not raise any concerns about the violation of the marginal distribution law, because also for the vessels of water, the detection events of the water correlations are space-like separated. So why are the quantum foundations physicists worried about the violation of the marginal distribution law, and whence their claims of it implying possible signaling faster than light, and a violation of causality? There is another, complex and non-direct mechanism that opens a way of possible signaling by means of this type of correlations. 

For a better understanding of this complex mechanism, we introduce the well-known protagonists Alice and Bob, always used in quantum foundations when this type of situation is analyzed. In our case, Alice is at the left reference vessel and Bob at the right one. Let us see how Bob could send a signal to Alice by means of the water correlations and this complex non-direct mechanism. If Bob decides to `constantly execute the siphon experiment, for example for a whole day', then Alice can note that during this day, the statistics at her side is of a certain type. Indeed, Alice finds for her siphon experiment $p(A_1,B_1)+p(A_1,B_2)={1 \over 2}$, and for her spoon experiment $p(A'1,B_1)+p(A'1,B_2)=1$. If Bob decides now the opposite, i.e. to constantly execute the spoon experiment, then Alice finds different statistics, namely $p(A_1,B'_1)+p(A_1,B'_2)=1$ for her siphon experiment, and $p(A'1,B'_1)+p(A'1,B'_2)=1$ for her spoon experiment. Hence, depending on what Bob does for a period of time, siphon or spoon, Alice will find the statistics of her siphon to change. Alice and Bob can agree beforehand to use this complex mechanism for signaling. For example, they could agree that Bob will use one of both options throughout the day, i.e. either siphon or spoon, to encode any message digitally, and send strains of bits to Alice, one a day. Alice would then only have to carefully observe her statistics over the course of each day, receiving in this way one bit of the message each day. It is easy to understand that this non-direct mechanism of signaling is always possible once the marginal distribution law is violated. Since the detection events for the vessels of water are space-like separated, would this then mean that signaling faster than light is possible with the water correlations? Of course not. If Bob changes the regime of his measurement from siphon to spoon and back, it takes time for this regime change to be reflected in AliceÕs statistics, which is why, with the water correlations, no violation of relativity can be produced, even if the detection events on both sides are space-like separated events. The reason that the majority of quantum foundation physicists believe that in the case of micro-physics, e.g. correlated photons, this way of signaling provokes a violation of relativity, if the marginal distribution law is violated, is that they imagine two hypotheses to be true: (a) the collapse is instantaneous, and (b) due to this instantaneous collapse, the statistics at AliceÕs side changes instantaneously when Bob changes his regime. If hypothesis (b) is true, then it is enough for Alice and Bob to have a distance of some light days between them, such that Bob can send signals faster than light to Alice, and violate relativity. But, does hypothesis (b) follow from hypothesis (a), which is what most quantum foundations physicists believe? The vessels of water example can show us that this is not necessarily so. If the instantaneous collapse happens at the detectors, as a consequence of the measurements affecting the entangled state (of the two photons), then the Ôtime involved in this stateÕ also plays a role in the messaging attempted by Bob. This Ôtime involved in the stateÕ is the time needed for the photons to fly half the distance between Alice and Bob, which is half the amount of days they are separated by light days. This will slow down BobÕs messaging possibly exactly so that signaling faster than light is excluded. Experiments can be envisaged to test the above, most probably revealing interesting new knowledge about the mechanism of collapse in the physical realm. Let us mention, without analyzing this here, that the correlation experiments in the physical realm do produce violations of the marginal distribution law, although to date no explanation and hardly any analysis has been given for this (Adenier, Khrennikov \& Yu, 2006).

4) We mention another macroscopic non local system we proposed in the past, giving rise to exactly the same correlation data like the ones produced by quantum spin 1/2 entangled particles (Aerts, 1991), where it can be seen that it is the presence of symmetry which gives rise to the marginal distribution law to be valid in this example. To illustrate this aspect within quantum cognition, we present a second example, which is a variation on our first violation of Bell inequalities example, using the concept {\it Cat}, and two of its exemplars {\it Glimmer} and {\it Inkling}, the names of two brother cats that lived in our research center at that time (Aerts et al., 2000). This example is also inspired by the first macroscopic non-local box example worked out by Sven Aerts already in 2005, by using a breakable elastic and well defined experiments on this elastic (Aerts, 2005).

We consider the concept {\it Cat} in its abstract state. The experiments that we introduce consist in realizing physical contexts that influence the collapse of the concept {\it Cat} to one of its exemplars {\it Glimmer} or {\it Inkling}, inside the mind of a person being confronted with the physical contexts. It is a `gedanken experiment', in the sense that we put forward plausible outcomes for it, taking into account the nature of the physical contexts, and Liane, the owner of both cats, playing the role of the person. One aspect of the physical context is that we supposed that our secretary would put a collar with a little bell around the necks of both cats, and sometimes not, hence with probability equal to ${1 \over 2}$ for both situations. Experiment $e_{A}$ consists in `{\it Glimmer} appearing in front of Liane as a physical context'. We consider outcome `1' to occur if Liane thinks of {\it Glimmer} and there is a bell, or if she thinks of {\it Inkling} and there is no bell, while outcome `2' occurs if she thinks of {\it Inkling} and there is a bell, or if she thinks of {\it Glimmer} and there is no bell. Experiment $e_{B}$ consists in `{\it Inkling} appearing in front of Liane as a physical context'. We consider outcome `1' to occur if Liane thinks of {\it Inkling} and there is a bell, or if she thinks of {\it Glimmer} and there is no bell, while outcome `2' occurs if she thinks of {\it Glimmer} and there is a bell, or if she thinks of {\it Inkling} and there is no bell. 
Experiment $e_{A'}$ consists in `{\it Inkling} appearing in front of Liane as a physical context', and outcome `1' occurs if he wears a bell, and outcome `2', if not. Experiment $e_{B'}$ consists in `{\it Glimmer} appearing in front of Liane as a physical context', and outcome `1' occurs if he wears a bell, outcome `2', if not.
Coincidence experiment $e_{AB}$ consists in both cats showing up as a physical context. Because of the symmetry of the situation, it is plausible to suppose probability ${1 \over 2}$ that Liane thinks of {\it Glimmer} and probability ${1 \over 2}$ that she thinks of {\it Inkling}, however, they are mutually exclusive. Also, since both cats wear bells or do not wear bells, $e_{AB}$ produces strict anti-correlation, probability ${1 \over 2}$ for outcome `12' and probability ${1 \over 2}$ for outcome `21'. Hence $p(A_1, B_2)=p(A_2, B_1)={1 \over 2}$ and $p(A_1, B_1)=p(A_2, B_2)=0$, which gives $E(A,B)=-1$. 
Experiment $e_{AB'}$ consists in {\it Glimmer} showing up as a physical context. This gives rise to a perfect correlation, outcome `11' or outcome `22', depending on whether {\it Glimmer} wears a bell or not, hence both with probability ${1 \over 2}$. As a consequence, we have $p(A_1,B'_1)=p(A_2,B'_2)={1 \over 2}$ and $p(A_1,B'_2)=p(A_2,B'_1)=0$, and $E(A,B')=+1$. Experiment $e_{A'B}$ consists in {\it Inkling} showing up as a physical context, again giving rise to a perfect correlation, outcome `11' or outcome `22', depending on whether {\it Inkling} wears a bell or not, hence both with probability ${1 \over 2}$. This gives $p(A'_1,B_1)=p(A'_2,B_2)={1 \over 2}$ and $p(A'_1,B_2)=p(A'_2,B_1)=0$ and $E(A',B)=+1$. Experiment $e_{A'B'}$ consists in both cats showing up as a physical context, giving rise to a perfect correlation, outcome `11' or outcome `22', depending on whether both wear bells or not, hence both with probability ${1 \over 2}$. This gives $p(A'_1,B'_1)=p(A'_2,B'_2)={1 \over 2}$ and $p(A'_1,B'_2)=p(A'_2,B'_1)=0$ and $E(A',B')=+1$.
For the CHSH inequality we find $E(A',B')+E(A',B)+E(A,B')-E(A,B)=4$. The marginal distribution law is satisfied in this example, because $p(A_1,B_1)+p(A_1,B_2)=p(A_1,B'_1)+p(A_1,B'_2)={1 \over 2}$. It is easy to check all the other formulas of the marginal distribution law. The above is a realization of what quantum foundations physicists call a `non local box'. It is amusing to note that since the example for each coincidence experiment only takes `one decision of Liane', the correlations are instantaneous.

With respect to Hampton (2013), we wish to make a first very short remark. If we have given the impression of treating graded membership and typicality as the same, we have certainly not meant to do so, because we are well aware of the difference. It is even one of the problems we want to investigate in the future, `how these different variables can be retrieved from what we believe to be the fundamental item, namely the probability linked to the decision that is involved in any measurement (e.g. the decisions of the subjects participating in the experiment). Hampton (2013) stimulates us to make more connections with existing models, such as his Composite Prototype model, and we could not agree more. In fact, it has been our aim for quite some time to do so -- in the experiment presented in Aerts \& Gabora (2005), we tested a whole collection of features (properties) of the considered concepts, and their representations in Hilbert space was finally left out because it would have made the article too long, Also here, we will be able only to give a very brief outline of how our quantum approach deals with features. In Aerts, Gabora \& Sozzo (2013), we sketch in which way our approach can be seen as a contextual prototype theory. Following on this, features would be represented as closed subspaces of the complex Hilbert space, and the weight of each feature with respect to the state of a concept, would be given by the quantum probability of collapse of this state on the subspace representing the feature. Hampton (2013) identifies `extension' and `intension' with respect to concepts, and classifies our approach as intensional. However, since we have been inspired by an existing formalism modeling quantum entities in the physical realm, our approach has been from the start to consider concepts as entities in states, expressing in this way the contextual behavior of quantities such as typicality and membership weight. Hence, there is no simple connection with `extension' and `intension', aspects introduced customarily in philosophical reflections about `meaning'. The quantum formalism applied to cognition gives another, potentially richer way of looking at meaning, coping with quite some of the problems expressed for example in Putnam (1975), with respect to the customary `extension' and `intension' based view. However, we definitely intend to engage in investigations of these matters and a concrete comparison with Hampton's Composite Prototype model could be a good start.

\bigskip
\noindent
{\bf References}

\small
\medskip
\noindent
Aerts, D. (1982). Example of a macroscopical situation that violates Bell inequalities. {\it Lettere al Nuovo Cimento}, {\bf 34}, pp. 107-111.

\noindent
Aerts, D. (1991). A mechanistic classical laboratory situation violating the Bell inequalities with $2\sqrt{2}$, exactly `in the same way' as its violations by the EPR experiments. {\it Helvetica Physica Acta}, {\bf 64}, pp. 1-23.

\noindent
Aerts, D. (2009). Quantum structure in cognition. {\it Journal of Mathematical Psychology}, {\bf 53}, pp. 314-348.

\noindent
Aerts, D., Aerts, S., Broekaert, J. and Gabora, L. (2000). The violation of Bell inequalities in the macroworld. {\it Foundations of Physics}, {\bf 30}, pp. 1387-1414. 

\noindent
Aerts, D. and Gabora, L. (2005). A theory of concepts and their combinations I\&II. {\it Kybernetes}, {\bf 34}, pp. 167-191; pp. 192-221.

\noindent
Aerts, D., Gabora, L. and Sozzo, S. (2013). Concepts and their dynamics: A quantum-theoretic modeling of human thought. {\it Topics in Cognitive Science}.

\noindent
Aerts, D. and Sozzo, S. (2011). Quantum structure in cognition: Why and how concepts are entangled. {\it Quantum Interaction. Lecture Notes in Computer Science}, {\bf 7052}, pp. 116-127.

\noindent
Aerts, D. \& Sozzo, S. (2013a). Quantum entanglement in concept combinations. Accepted for publication in {\it International Journal of Theoretical Physics}.

\noindent
Aerts, D. \& Sozzo, S. (2013b). Entanglement zoo I: Foundational and structural aspects. To be presented at {\it Quantum Interaction 2013, Leicester, UK}.

\noindent
Aerts, S. (2005). A realistic device that simulates the non-local PR box without communication. \url{http://arxiv.org/abs/quant-ph/0504171}. 

\noindent
Adenier, G., Khrennikov, A. Yu. (2006). Anomalies in EPR-Bell
experiments. In Adenier, G., Khrennikov, A. Yu., Nieuwenhuizen, T. (Eds.), {\it Quantum Theory: Reconsideration of Foundations 3}, pp. 283-293. New York: AIP.

\noindent
Dzhafarov, E.N., \& Kujala, J.V. (2013). On Selective Influences, Marginal Selectivity, and Bell/CHSH inequalities. {\it Topics in Cognitive Science}.

\noindent
Hampton, J. (2013). Conceptual combination: extension and intension: Commentary on Aerts, Gabora and Sozzo. {\it Topics in Cognitive Science}.

\noindent 
Putnam, H. (1975). The meaning of `meaning'. {\it Minnesota Studies in the Philosophy of Science}, {\bf 7}, pp. 131-193.
\end{document}